# Pragmatic Frames Evoked by Gestures: A FrameNet Brasil Approach to Multimodality in Turn Organization

Helen de Andrade Abreu [a]
https://orcid.org/0000-0003-1228-7144

Tiago Timponi Torrent [ab]
https://orcid.org/0000-0001-5373-2297

Ely Edison da Silva Matos[a]
https://orcid.org/0000-0002-9464-9315

[a]FrameNet Brasil Lab, Federal University of Juiz de Fora (UFJF), Juiz de Fora, Minas Gerais, Brazil
[b]National Council for Scientific and Technological Development (CNPq)

## ABSTRACT

This paper proposes a framework for modeling multimodal conversational turn organization via the proposition of correlations between language and interactive gestures, based on analysis as to how pragmatic frames are conceptualized and evoked by communicators. As a means to provide evidence for the analysis, we developed an annotation methodology to enrich a multimodal dataset (annotated for semantic frames) with pragmatic frames modeling conversational turn organization. Although conversational turn organization has been studied by researchers from diverse fields, the specific strategies, especially gestures used by communicators, had not yet been encoded in a dataset that can be used for machine learning. To fill this gap, we enriched the Frame² dataset with annotations of gestures used for turn organization. The Frame² dataset features 10 episodes from the Brazilian TV series Pedro Pelo Mundo annotated for semantic frames evoked in both video and text. This dataset allowed us to closely observe how communicators use interactive gestures outside a laboratory, in settings, to our knowledge, not previously recorded in related literature. Our results have confirmed that communicators involved in face-to-face conversation make use of gestures as a tool for passing, taking and keeping conversational turns, and also revealed variations of some gestures that had not been documented before. We propose that the use of these gestures arises from the conceptualization of pragmatic frames, involving mental spaces, blending and conceptual metaphors. In addition, our data demonstrate that the annotation of pragmatic frames contributes to a deeper understanding of human cognition and language.

## 1. INTRODUCTION

The study of meaning in Cognitive Linguistics in general and in Frame Semantics in particular has always denied a strict separation between semantics and pragmatics, treating them as a continuum grounded on experience. Nonetheless, the computational implementation of Frame Semantics, namely FrameNet, has devoted most of its time and attention to modeling frames encoding meaning that is more content-oriented and less procedural. In other words, most of the frames defined in a FrameNet model events—such as buying, selling, moving—, attributes—color, shape, size—, entities—people, animals, artifacts—and relations—part-whole, addition, concession. In this paper, we turn our attention to frames modeling phenomena that traditionally fall under the realm of pragmatics. In specific, we look into turn organization frames and to their evocation by co-speech gesture.

Semantics and pragmatics are traditionally treated separately, even when they are seen as complementary aspects of language. However, as Marmaridou (2000) explains, Experiential Realism and Cognitive Semantics offer the theoretical basis for a view of both semantics and pragmatics as deriving from the same source and being closely related in a way that could not be totally understood before through other philosophical approaches. Cognitive Semantics is based on a number of tenets derived from the philosophical framework of Experiential Realism (Lakoff, 1987; Lakoff and Johnson, 1999). Among these tenets is the view of language as part of human general cognition and that language is also developed in society; in other words, language has a biological as well as a societal basis. As Marmaridou (2000) demonstrates, different pragmatic phenomena—speech acts, implicature, deixis and presupposition—can be analysed through Experiential Realism, using Cognitive Semantics principles such as Mental Spaces Theory (Fauconnier, 1994, 1997; Fauconnier and Turner, 2002), frames (Fillmore 1975, 1982), MCIs (Lakoff, 1987), categorization (Rosch, 1973; Rosch and Lloyde, 1978) and others. The potential for this approach in pragmatic research, however, is not exhausted, as Marmaridou (2000, p. 11) points out.

Aiming to contribute to filling part of this gap, this paper starts with a brief explanation of frames and frame semantic theory, and a presentation of FrameNet—specially FrameNet Brasil—followed by the Frame² dataset. It proceeds to demonstrate how a cognitive linguistics approach enriches pragmatics studies, and the importance of developing annotation of pragmatic frames. It then provides a brief overview of some basic concepts of cognitive linguistics that are necessary for understanding our analyses and discussion. Next, this paper briefly presents the subjects of conversation analysis, turn organization and interactive gestures. Section 2 presents our methodology; section 3 details our analyses; and section 4, our discussion, based on our findings.

## 1.1. FRAMES, FRAME SEMANTICS AND FRAMENETS

Frame Semantics was first developed by Charles Fillmore (1975; 1982; 1985; 2008), who used the term 'frame' to refer to "a system of concepts related in such a way that to understand any one of them you have to understand the whole structure in which it fits (…)" (Fillmore, 1982:111). According to the author, every time an element of that structure is evoked, the other elements in the structure are made available. So, for example, for a person to understand the sentence "I bought this car from a friend", it is necessary for that person to have, as part of their conceptual system, the structure which involves sellers, buyers, goods, etc. This particular structure was called the "commercial event" frame by Fillmore (1982), and it includes, as some of its frame elements, sellers, buyers and goods.

Fillmore created, in 1997, the FrameNet initiative (Fillmore et al., 2003a; 2003b) at the International Computer Science Institute (ICSI) in Berkeley, California. FrameNet's aim was to build a lexicographic database whose analysis could be attested by the annotation of usage data taken from corpora of the English language, which could be useful both for the advancement of research in linguistics and for machine learning tasks.

Although Fillmore was primarily concerned with frames that were evoked through language, he recognized that frames are part of the foundations of cognition in general. In his first work on the subject (Fillmore, 1975), he treated frames and scenes as similar phenomena—the difference between the two arising from the fact that while frames were lexically activated, scenes were activated through other cognitive means. Nonetheless, Fillmore (2008) later decided that, in fact, both frames and scenes consist of a single phenomenon that could be referred to by the same term, i.e., 'frame'.

After Berkeley FrameNet was established, other FrameNets were created, developing studies in languages such as German, Japanese and Spanish (Subirats-Rüggeberg; You; Liu, 2005; Boas; Ziem, 2018; Gruzitis et al. 2018; Ohara et al., 2018; Hahm et al. 2020). Among them is FrameNet Brasil, developed at the FrameNet Brasil Laboratory, at the Federal University of Juiz de Fora since 2008, to work with Brazilian Portuguese.

## 1.2 FRAMENET BRASIL AND FRAME²

FrameNet Brasil started as a lexicographic database adapted to the particularities of Brazilian Portuguese. The FN-Br WebTool (Torrent & Elsworth, 2013; Torrent et al., 2024) is the tool developed for the annotation of semantic frames. The process of annotation involves the analysis of each lexical unit (a frame-evoking word or term) present in a sentence. Using the WebTool, annotators manually choose each lexical unit (LU) and then mark each of the frame elements belonging to that specific frame which are present in the sentence.

Fig. 1 shows an example of how a frame is seen on FrameNet Brasil database. This is the `Possession` frame, containing the frame definition and its frame elements (together with their own definitions), divided into core, peripheral and extra-thematic, according to their importance for the expression of the frame in a sentence.

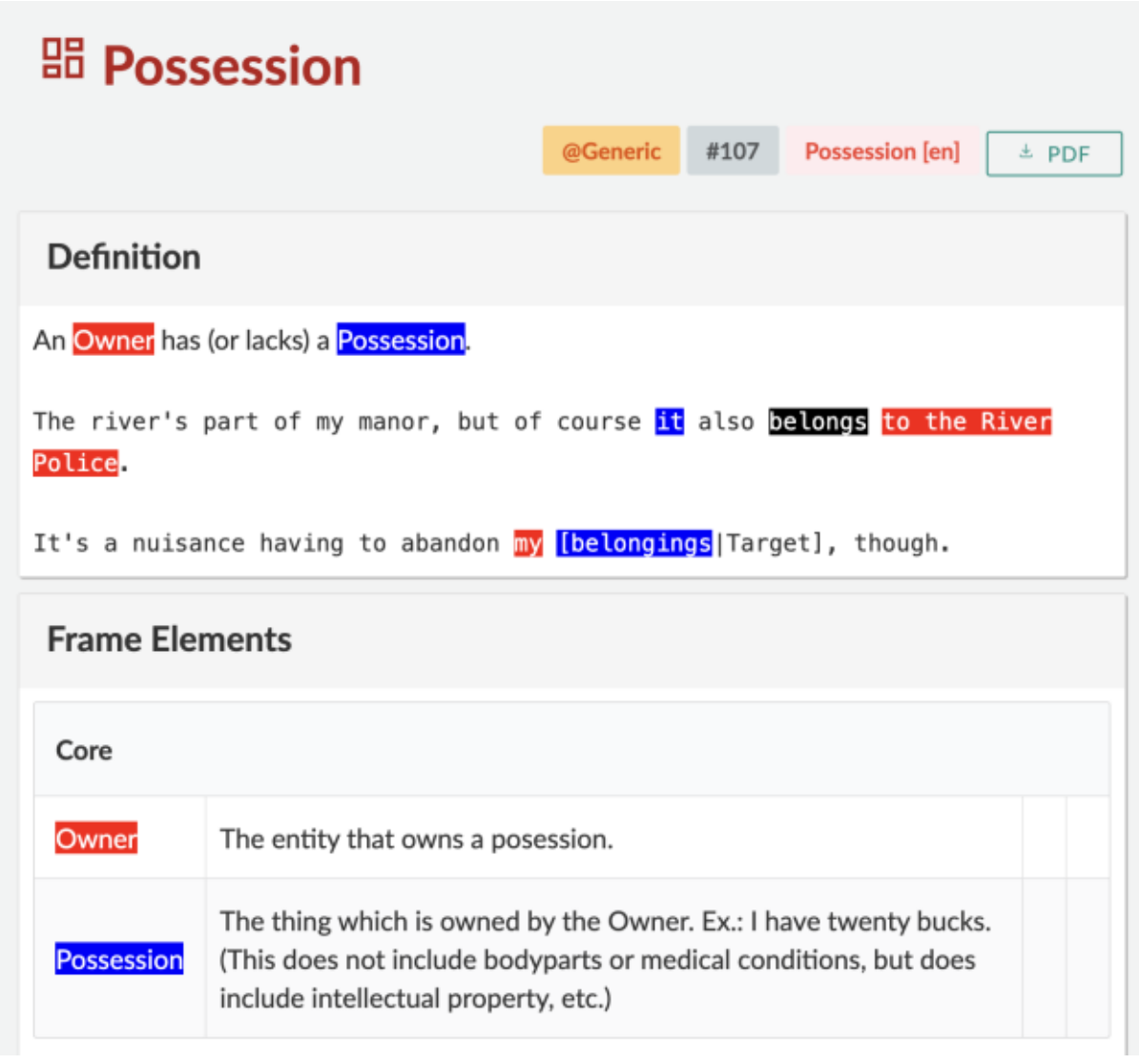

Figure 1 - `Possession` frame

Fig. 2 shows the sentence "Scotland always had its own culture", and the way it was annotated for its lexical units: 'always' and 'have'. As can be observed, the LU 'always' evokes the `Frequency` frame; and the LU 'have' evokes the `Possession` frame. Both frames have frame elements expressed in the sentences, labeled in different colors. The LU 'have', for example, evokes the Possession frame, which contains two frame elements present in the sentence: Owner (Scotland), marked in red, and Possession (its own culture), marked in blue.

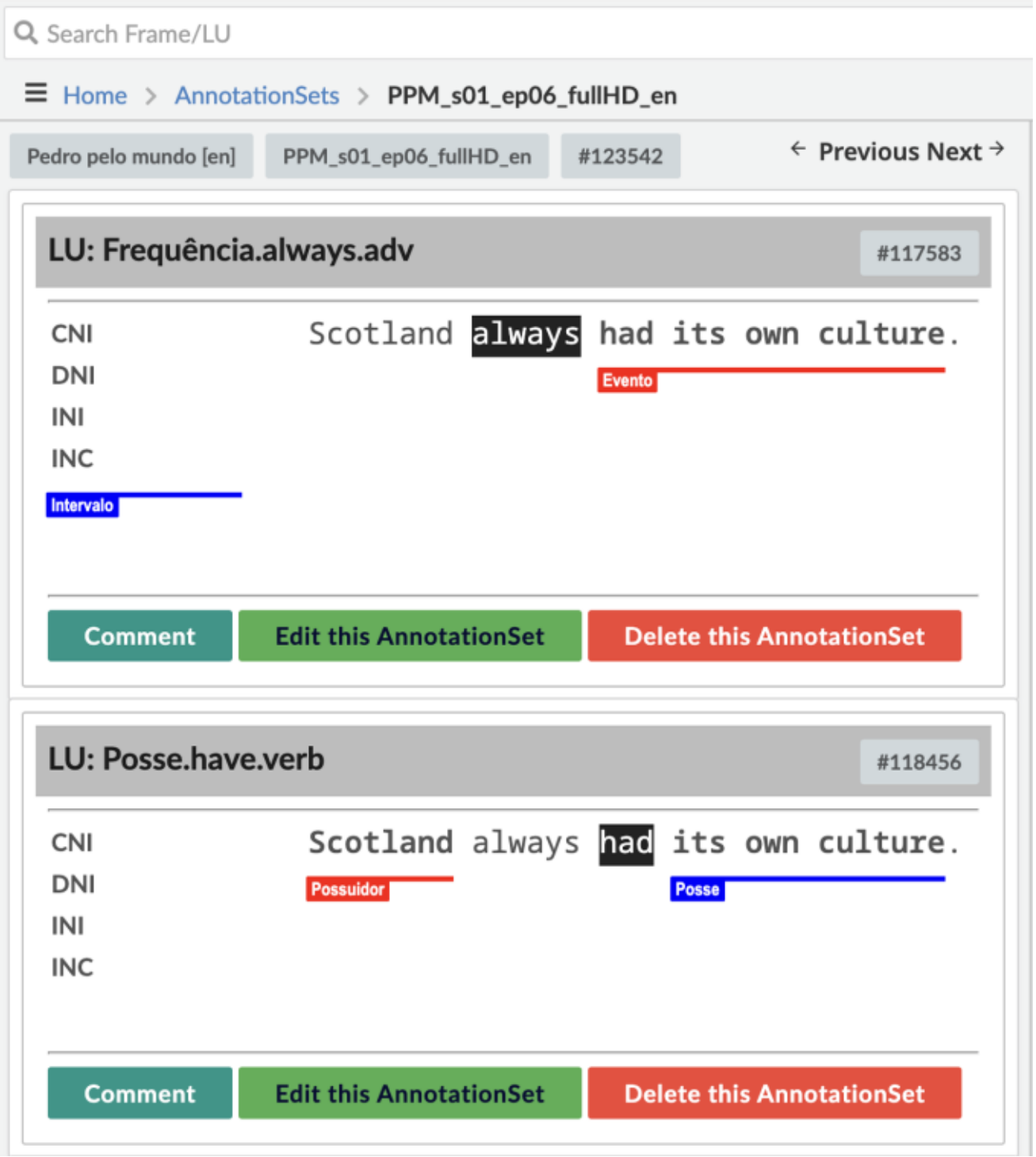

Figure 2 - Annotation set

Even though it started as a lexical database, FrameNet Brasil has been developed into a multimodal database (Belcavello et al., 2022; Viridiano et al., 2022), and it is now capable of working with frames not only in textual corpora, but also in static and dynamic images.

Frame² (pronounced "frame squared"), a multimodal dataset (Belcavello et al., 2024) within FrameNet Brasil, was developed using the television series Pedro Pelo Mundo[1]. All the spoken language in these ten episodes, as well as all the subtitles present (considered together as text) were fully annotated, generating a total of 16,458 annotation sets. Besides that, the WebTool provided the resources necessary so that the visual part of these episodes could also be annotated for objects and the frames they evoked. As seen in section 1.1, frames are part of the basis of our cognition and, therefore, can be activated by other means rather than language. More importantly, as Belcavello et al. (2020) demonstrated, images and text can produce meaning as a result of functioning together. A total of 6,841 visual objects were also annotated in the ten episodes (Belcavello et al., 2024), proving that FrameNet Brasil was enriched by the multimodal approach.

## 1.3 PRAGMATIC FRAMES

Czulo, Ziem and Torrent (2020) argue, in their paper *Beyond Lexical Semantics*, in favor of expanding the domain of frames that traditionally make up the FrameNet database to the discursive domain with pragmatic frames. The authors depart from the results obtained through the Global FrameNet Shared Annotation Task (Torrent et al., 2018) and demonstrate, through the comparison of examples from English, German and Brazilian Portuguese, that the annotation of pragmatic properties is productive cross-linguistically. In other words, although the forms may be different in each of the languages analyzed, the pragmatic meaning may be correlated. One of the examples provided by the authors is the greeting 'bom dia' (in Brazilian Portuguese), 'good morning' (in English) and 'gutten morgen' (in German). In the shared annotation task, Brazilian and American annotators chose not to annotate the words that form the greeting, recognizing the pragmatic meaning that until then could not be annotated in FrameNet. The German annotators, however, chose to annotate each word separately, as seen below (Czulo, Ziem and Torrent, 2020):

(1) [Gutten $_{\text{Desirability}}$] [Morgen $_{\text{Calendric_unit}}$]
*good* *morning*

As the authors explain, the pragmatic meaning, a greeting, uttered by people who are meeting for the first time in a certain period of the day, cannot be expressed simply by the frames evoked by each of the lexical units separately (`Desirability` and `Calendric_unit`). For this reason, Czulo, Ziem and Torrent (2020) claim that there is a need for FrameNet to add a pragmatic layer to the ones already present on WebTool annotations.

The pragmatic approach is in line with the ideas of Fillmore himself, who stated in an interview with József Andor (2010), "For me semantics proper is a study of the relation between matters of *linguistic form* and a community's *conventions* that shape native speakers' interpretations of uses of such form." (Andor, 2010) (italics in the original). This approach is also in conformity with the notion of frame developed by Erving Goffman in his book Frame Analysis (Goffman, [1974] 1986), in which the author states:

> I assume that definitions of a situation are built up in accordance with principles of organization which govern events—at least social ones—and our subjective involvement in them; frame is the word I use to refer to such of these basic elements as I am able to identify. (Goffman, [1974] 1986, pp. 10-11).

## 1.4 MENTAL SPACES, BLENDING AND BASIC COMMUNICATIVE SPACES NETWORK

The explanation of how pragmatic frames are evoked involves one of the most fundamental theories within Cognitive Linguistics—the Mental Spaces Theory (Fauconnier, [1985] 1994, 1997; Fauconnnier and Turner, 2002). As Fauconnier and Turner (2002, p. 40) explain, mental spaces are conceptual domains that are constructed locally and structure thoughts and communication. They are generally structured by frames, encyclopedic knowledge, or specific knowledge, such as knowledge about a fact that occurred previously. This way, if a person talks about a purchase they made (to continue with the example used before), a mental space is built which contains as its elements the person who made the purchase (the utterer, in this case), the product purchased and as many elements as are introduced in the information shared. Other mental spaces are built as the conversation progresses, or the same mental space may be accessed as necessity demands.

A very important characteristic of mental spaces is that they can be connected through mappings[2]. This way, elements in a mental space can be related to other elements in other spaces, in a complex conceptual system[3].

[1] FrameNet acquired the rights to use the ten episodes of the first season of this series for research purposes. The example in Fig. 1 was taken from this dataset.

[2] 'Mapping' is a term taken from mathematics and, according to Fauconnier (1997, p. 1), "is a correspondence between two sets that assigns to each element in the first a counterpart in the second".

[3] Although we use in our work a view from the standpoint of Cognitive Linguistics, it is noteworthy that the same phenomena may be seen from a neuroscience point of view. In the case of mental spaces and mappings, according to Fauconnier and Turner (2002, p. 40): "In the neural interpretation of these cognitive processes,

Mappings are what is behind the understanding of a sentence like "I bought a car from a friend". As we mentioned before, if a person talks about a purchase they made, a mental space is built containing the elements mentioned. To be more precise, this mental space is connected to the mental space organized by the commercial event frame mentioned in section 1.1. People are capable of producing this kind of sentence, and of understanding it upon hearing or reading it, because mappings occur between the elements of one mental space containing the roles (frame elements) of buyer, seller, goods, etc, and the other mental space containing the people and the objects that will become, through this process, connected to these roles. This way, the utterer ("I") is mapped to the buyer, "a friend" is mapped to the seller, and "a car" is mapped to the goods, generating the conceptualization of what happened. This conceptualization needs to happen, first, in the mind of the utterer, so that they can communicate the idea "I bought a car from a friend". Once the comprehender hears or reads this sentence, the same conceptual processes happen, so that they can understand the meaning of what was communicated.

Conceptual integration, or blending, is a more complex system of mappings through which certain mental spaces (which become input spaces) are integrated is such a way as to generate a new mental space (the blending space) containing elements of the input spaces, but forming a gestalt (the blend itself). Blending is a process responsible for much of human conceptualization, including the use of metaphors, which are much more pervasive in human cognition than previously understood[4]. More on how blending works in the conceptualization of pragmatic frames will be presented in section 4.

The Basic Communicative Spaces Network is a mental spaces approach (Sanders et al. 2009; Sweetser, 1990; Dancygier and Sweetser, 2005; and Ferrari and Sweetser, 2012) according to which, whenever communication occurs, a certain network of mental spaces is always available. One part of this network is an expansion of the concept of *ground* presented by Langacker (1990): it contains a ground base space which is the mental representation, on the utterer's part, of utterer[5] (speaker, originally), comprehender (hearer, originally), moment and place of the communicative act. Moreover, the ground network also includes, as explained by Ferrari and Sweetser (2012, p. 50), an epistemic space (or more) for the communicators' mental processes and their mental states; a speech act space for the kind of interaction that the communicators are engaged in; and a metalinguistic space for reference to language use itself.

The Basic Communicative Spaces Network (BCSN) may also include a content space of what is being communicated at a certain moment—which can only be accessed through the ground network. However, in certain cases, what is said is understood directly via the ground network itself. Ferrari and Sweetser (2012, p. 56-58) provide an example of each situation. One of them is the utterance *God be with you*, which expresses the utterer's wish that the comprehender be in the presence of God upon departure. In this case, the content space contains a content base of the utterer's communication and a wish space of the idea that God be with the comprehender. This content can only be conceptualized through the ground base space (containing the participants in the conversation) and the speech act space (containing the context of the conversation). On the other hand, the second example, the utterance *Good-bye*, is understood via the ground network itself, because the expression *good-bye* has lost its previous meaning ('God be with you'), becoming solely a marker of departure. It is understood through the ground base space, containing the mental representation of utterer, comprehender, place and time of the communicative act, and the speech act space of the parting between the utterer and the comprehender. The wish is not expressed anymore, and this way does not generate a content space.

As will be discussed in section 5 of this paper, we understand that, in fact, processes of blending are involved in how the BCSNs work and also how BCSNs are involved in the conceptualization of pragmatic frames.

## 1.5 CONVERSATION ANALYSIS, TURN ORGANIZATION AND INTERACTIVE GESTURES

As our work involves the use of pragmatic frames evoked in conversational turn organization, it is fundamental to understand how face-to-face conversation is characterized and the rules and strategies adopted by communicators for turn taking.

In their foundational paper, Sacks, Schegloff, and Jefferson (1974) propose a model for how turn-taking in conversation is organized. As they defined, this model should be "context-free" and, at the same time, "context-sensitive"; that is, the model itself should not depend on context, such as people involved, place and time when the conversation took place, but it should be a basis on which different contexts could work, with adaptations to their different needs.

The authors list a series of fourteen characteristics of any conversation, and they also propose a set of specific rules which form the basis for turn construction. Turn-tranfers occur at what the authors name transition-relevance

---

mental spaces are sets of activated neuronal assemblies, and lines between elements [as presented in graphic form] correspond to coactivation-bindings of a certain kind".

[4] For more on metaphors, and specifically conceptual metaphors, we suggest Lakoff and Johnson ([1980] 2003).

[5] We decided to depart from the traditional terms, using 'utterer' instead of 'speaker', 'comprehender' instead of 'hearer' and 'communicators' instead of 'interlocutors'.

places—the moment at conversation when the role of utterer (originally 'speaker') is taken by another party (or, as the rules provide for, retaken by the same party who had already been speaking). At this moment, the turn may also be offered by the utterer to a selected party.

The authors defend the idea that conversation is the main application for language, and therefore, as turn-taking is the principal characteristic of conversation, language must have syntactic rules for turn-taking. They also mention the fact that prosody has an important role in indicating that a turn is coming to its end, thus preparing communicators for the possibility of taking the role of next utterer. It is noteworthy that they also mention the fact that, when an utterer chooses to select the next utterer, this may be attained through the use of gaze direction. It demonstrates that Sacks, Schegloff, and Jefferson (1974) recognize the existence of some form of gesture used in turn management and, therefore, they recognize multimodality in face-to-face conversation, even though the authors themselves do not use this term, nor do they make it a central subject of their paper.

Other authors (Ostermann & Garcez, 2021; Bavelas et al., 1992; Bavelas 2022) have added to what nowadays is an extensive body of work on conversation analysis. However, the contribution of Sacks, Schegloff, and Jefferson (1974) remains relevant.

One of these authors, Bavelas (2022), explores the important issue of what constitutes face-to-face dialogue, which she defines as the first form of communication learned by infants, and the only one that is shared by all human societies and cultures all over the globe (Bavelas, 2022:1). The author lists seven characteristics of this mode of communication[6] (Bavelas, 2022: 3-5): first and foremost, the fact that in a face-to-face dialogue there are at least two people present—the utterer and the comprehender, who is not just a listener, but a person who will become an utterer, as the roles of utterer and comprehender change during the course of dialogue; in this type of communication, utterer and comprehender(s) are present in the same location, sharing a social context; as they are present in the same place, utterer and comprehender(s) are visually in contact with each other without any time lapse (as might occur during a video conference); in order to be considered conversation, it has to be spontaneous, and not rehearsed in any way; face-to-face dialogue makes use of words, as well as of gestures and facial expressions, making this kind of dialogue **multimodal**—it is important to note that these gestures and facial expressions happen at the same time as the verbal communication, and not in a linear way; there is an overlapping of communication on the part of the communicators, who can use gestures and facial expressions at the same time that the other person is speaking, and sometimes they may even talk at the same time or interrupt each other; words and gestures are "ephemeral"—they don't last, but communicators keep them in their minds and refer back to them, maintaining a continuation of ideas[7].

According to Bavelas (2022, p.13), "it is necessary to change the focus from individuals to their interaction; change the focus from nonverbal communication (or 'body language') to co-speech gestures". In studying how interactive gestures derive from and evoke pragmatic frames our work contributes to this change of focus.

Bavelas et al. (1992) were the first authors to establish the importance of differentiating between two types of co-speech gestures—topic and interactive gestures. As they proposed, topic gestures are those used to illustrate the topic of conversation—for example, a person talking about someone walking, while making a gesture with two fingers representing the action. On the other hand, interactive gestures refer specifically to the utterer's interlocutor—for example, passing the next turn of conversation to them, without making any reference to this gesture through words. According to the authors, "[interactive gestures] maintain involvement with the interlocutor without interrupting the verbal flow of discourse" (Bavelas et al., 1992:469).

Interactive gestures may be accompanied by a reference to the comprehender in the form of words, but most often they are the only mode of communication that makes this reference, while the words being uttered refer to the topic of conversation. In order to decide whether a gesture is interactive or not, the authors developed a specific procedure. First, the observer must check whether the gesture illustrates something that was said by the utterer (making it a topic gesture). If not, the observer should have in mind that:

> To be an interactive gesture, it must have a paraphrase that is both independent of the topic and addressed to the interlocutor. In addition, the form must be interactive, which means that the finger(s), thumb, or open palm(s) are oriented directly toward the other person at some point, however briefly. (Bavelas et al., 1992)

This way, face-to-face conversation is proved to be multimodal, with two different layers of communication being conveyed at the same time through verbal language and interactive gestures.

The authors demonstrated something of great importance to our work: the fact that interactive gestures were also used in turn organization during face-to-face conversation. Among the gestures depicted in Bavelas et al. (1992: 474-475), three of them are the most relevant for our present work: the gesture that is a request for help (palm upward, hand making a movement in circles in the air); the gesture that indicates request for keeping the turn (palm facing the listener, arm bent upwards); and the gesture that indicates turn offering (the speaker extends his/her arm in the direction of the listener, keeping his/her palm turned upwards, fingers slightly flexed, as if supporting an object). As the authors note, the "conduit metaphors", as named by McNeill and Levy (McNeill,

[6] Bavelas (2022, p. 3-5) makes it clear that this list "owes a great deal to Linell ([1982] 2005) and Clark (1996)"

[7] Which, as will be demonstrated in section 4, is in accordance with the Mental Spaces Theory.

1987; McNeill & Levy, 1982), are gestures that represent the delivery of information as if it were the delivery of a physical object. Turn of speech is generally "passed" the same way[8].

Interestingly, the authors point out that "(…) no two interactive gestures are exactly alike, but they do have recognizable common features" (Bavelas et al., 1992: 471). It is our interpretation that this can be explained according to the view of categorization presented by Rosch (1973) and Rosch & Lloyde (1978), according to which categorization is not conceptualized as an all or nothing situation. Rather, human cognition uses a prototype as a model at the center of a category, and other items are recognized as being part of the category according to how closely they resemble that prototype. An item that is too different from that prototype does not belong to the same category. This way, we propose that an interactive gesture may also have a prototype. In the case of the "passage of turn" gesture, for example, the most prototypical one is that which resembles the delivery of an object directly to the listener, as described above. As we will demonstrate in section 4, this prototype allows us to recognize variations of the same category of turn passing gestures.

According to Bavelas et al. (1992), considering previous literature on Broca's and Wernicke's aphasias (McNeill, 1985; Pedelty, 1987), one of the possible implications of their findings is that "social as well as semantic and syntactic aspects of language are 'hard-wired'" (Bavelas et al., 1992: 486). We notice that this is in accordance with an Experiential Realism and Cognitive Semantics view of language (Lakoff, 1987; Marmaridou, 2000).

It should be noted that Bavelas et al. (1992) make it clear that interactive gestures are not the only system used in turn organization, as there are other resources used by interlocutors that encompass syntax and intonation as well as facial expressions[9], which the authors also consider interactive. However, their specific findings on interactive gestures are very significant to our work.

## 2. METHODOLOGY

The corpus chosen for our work is the first season of the travel series Pedro Pelo Mundo, composed of ten episodes of twenty-three minutes each. In this series, Pedro Andrade visits cities around the world showing his viewers interesting places, trying the local cuisine, and also interviewing people who live in the location; among them, some Brazilians. The interviews are mostly in Brazilian Portuguese or in English. We chose this corpus having in mind the fact that FrameNet Brasil had already acquired permission for the use of these episodes for research, and also the fact that all the spoken part of each episode—as well as a host of static and dynamic images—had previously been annotated for semantic frames (Belcavello et al., 2020; Belcavello et al., 2022; Belcavello et al., 2024), as explained in section 1.2. This way, the different modes of annotation can be used in future developments of this work. To develop our work, the ten episodes of the series were transformed into a subdataset within Frame[2], named Pedro Pelo Mundo – Gestures.

We are aware of certain limitations caused by the format in which the series was filmed. For example, the camera was not always focused on the communicators, showing, instead, images that illustrated what was being discussed. Also, sometimes the scene was cut before possible gestures were performed. However, there are advantages to this format that benefited our work – we were able to observe forms of interactive gestures that had not been, to our knowledge, described in previous literature. For example, we observed a few turn passing gestures that were not exactly as described previously (as seen in 1.5 above). These observations only happened because the interlocutors were in a setting that could not be observed in a laboratory. These gestures will be shown and analyzed in the next section.

WebTool was the tool used for annotation of the interactive gestures used by Pedro Andrade and his interviewees during their conversations. To develop our research, it was necessary, first, to develop annotation of pragmatic frames on FrameNet Brasil. It is interesting to point out that German FrameNet and Constructicon was the first FrameNet to work on defining pragmatic frames (Ziem; Willich; Triesch, 2023). We used their model for the pragmatic frame `Communicative_context`, the first one developed on WebTool. Having the arguments of Czulo et al. (2020) as guidelines, we then developed the pragmatic frame `Greetings`, followed by the specific pragmatic frames associated with the turn organization gestures we expected to find in the corpus—`Organization_of_conversation` (which the following pragmatic frames would use), `Turn_passing`, `Turn_taking`, `Turn_keeping` and `Turn_confirmation`. We could develop more pragmatic frames related to turn organization, in case our observations proved it necessary. Once the pragmatic frames were developed, they were available for annotation.

Fig. 3 shows the `Organization_of_conversation` frame as seen on WebTool. The Frame-Frame Relations panel shows that this frame uses the `Communicative_context` frame and has, as subframes, the specific turn organization frames mentioned above. The Frame Elements panel in Fig. 3 shows Communicators, Comprehender and Utterer, which are shared with the other pragmatic frames related to turn organization.

8 Metaphors are an area of great interest in Cognitive Linguistics, and can be explained through processes of conceptual integration (blending), as will be demonstrated in section 4.

9 As pointed out by Sacks, Schegloff, and Jefferson (1974).

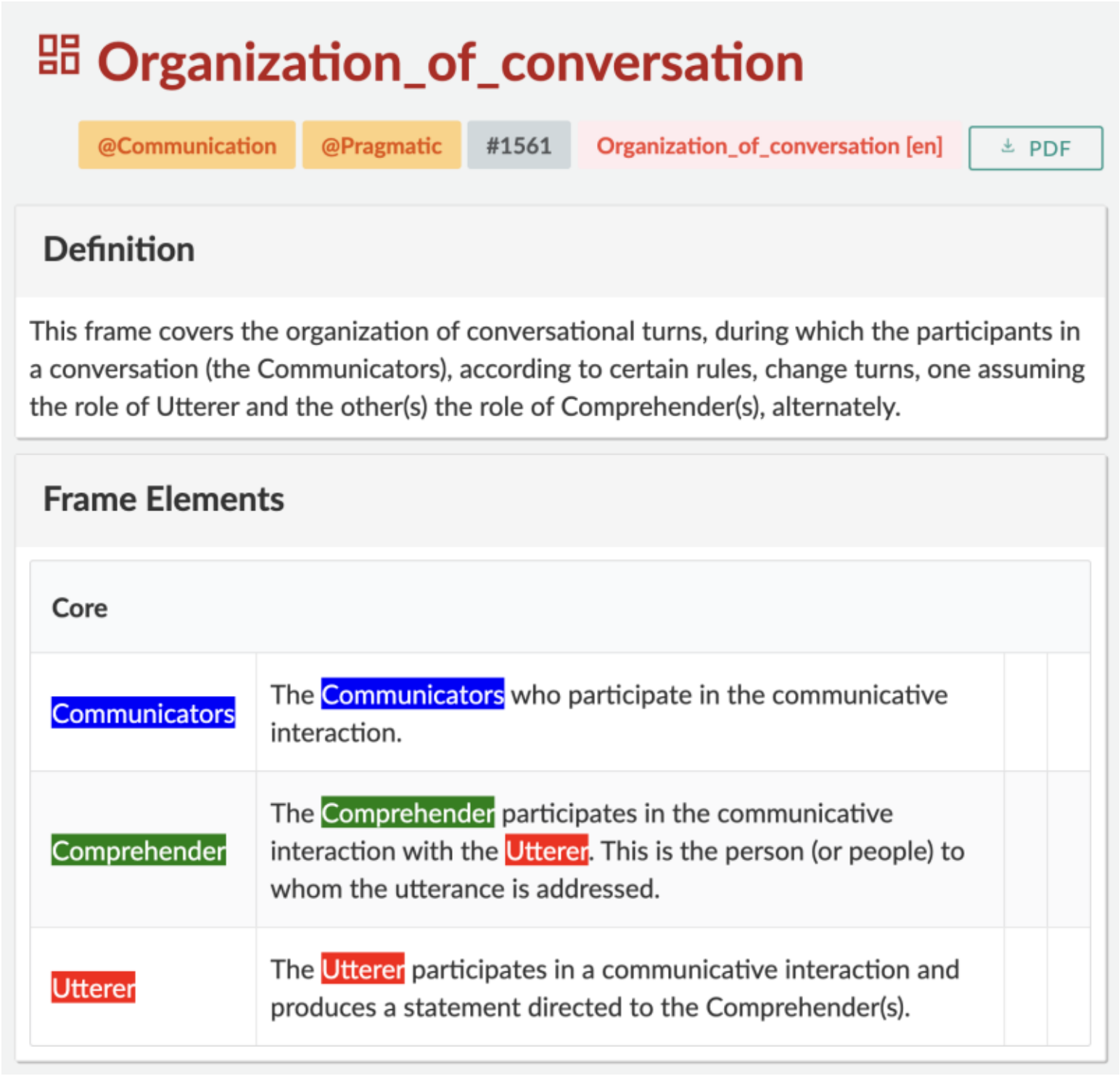

Figure 3 - `Organization_of_conversation` frame

As explained, the WebTool already had the tools for the annotation of dynamic images. These tools were then used for the annotation of each of the interactive gestures observed in the ten episodes of the first season of Pedro Pelo Mundo. For this task, we were guided by Sacks, Schegloff, and Jefferson (1974)'s work for the recognition of transition-relevance places and also by the parameters established by Bavelas et al. (1992) for the recognition of interactive gestures.

Fig. 4 shows a moment in episode 5, as seen on WebTool, in which Pedro Andrade uses the prototypical turn passing gesture as described in Bavelas et al. (1992) evoking the `Turn_passing` frame. This gesture, composed of hand and head movements together, was manually annotated with a bounding box for the hand and another for the head. Each bounding box marks Pedro Andrade's movements from the beginning of the gesture until the moment he finishes it or until the scene is cut. The WebTool still lacks the means to annotate both movements as only one unit; for this reason, the hand movement and the head movement had to be annotated separately. Fig. 4 shows both bounding boxes; however, in the panel on the upper right hand corner, only one object at a time can be selected. In this case, the object (CV name[10]) is *`Partes_do_corpo`: mão* ('`Body_parts`: hand'). The bottom right hand corner of the figure shows each movement annotated as a separate object (the number of the object corresponds to the number of its bounding box), the moment the movement starts, the moment it ends, its identification number, the frame for which it was annotated, the frame element and the CV name with its corresponding frame and lexical unit (LU). This procedure was used for each interactive gesture involved in turn organization observed in the episodes.

[10] 'CV name' stands for 'Computer Vision name'.

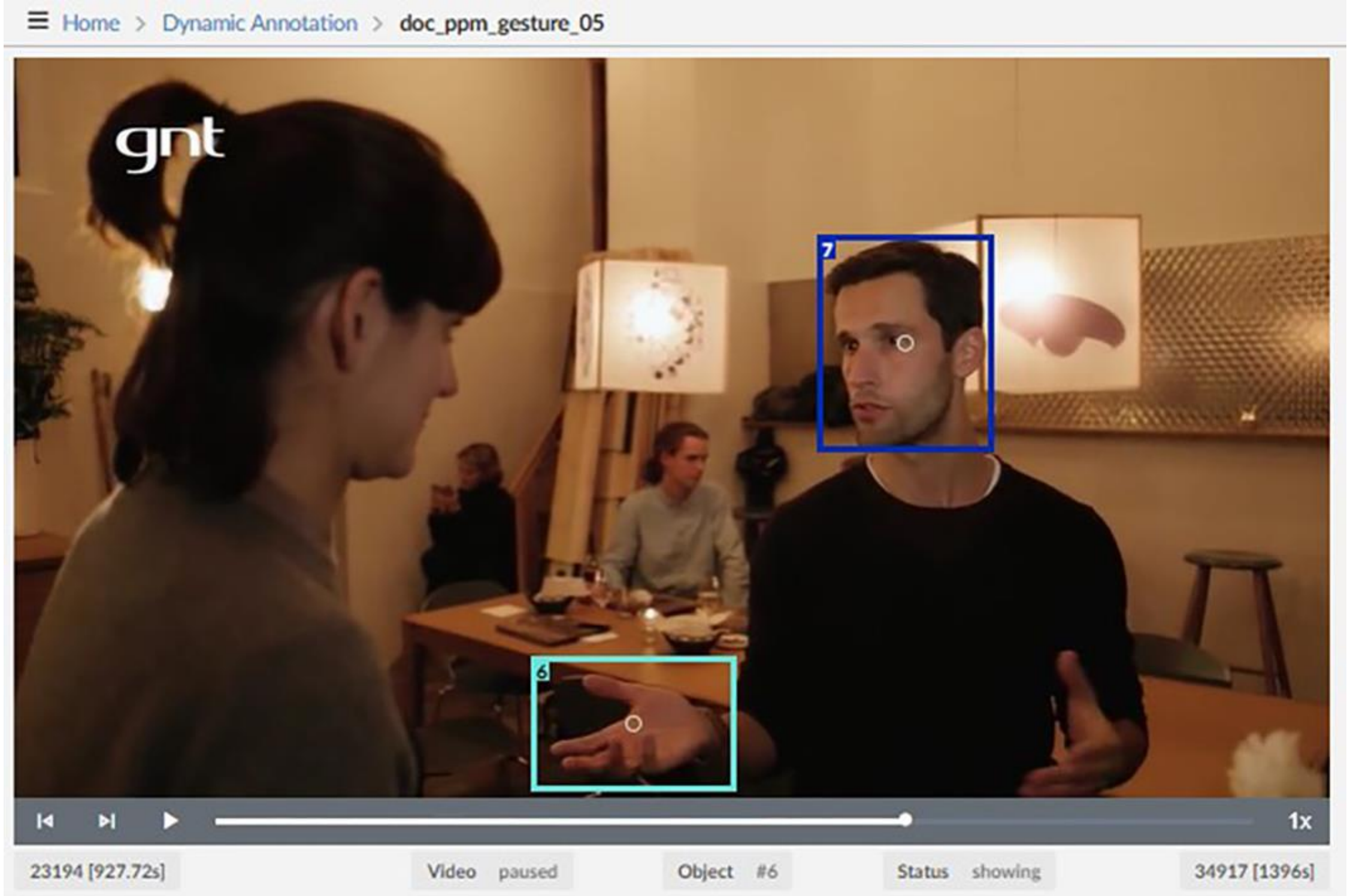

Figure 4 - Prototypical turn passing gesture

## 3. ANALYSIS

We were able to identify a total of 48 interactive gestures used for turn organization. Of this total, 30 evoke the `Turn_passing` frame, 16 evoke the `Turn_confirmation` frame, and 2 evoke the `Turn_taking` frame. We were not able, however, to identify the usage of any gestures that evoke the `Turn_keeping` frame.

We assume that the nature of our corpus affects the type and number of interactive gestures identified. As explained in section 2, the scenes observed were all part of interviews, which accounts for the high number of turn passing gestures. Also, these interviews are part of a series about traveling. Due to the characteristics of the series, Pedro Andrade encourages his interviewees to talk as much as possible, without interruption, about their experiences. For this reason, there are situations in which he confirmed through his interactive gestures that the interviewee should proceed talking (`Turn_confirmation` frame). There were only two instances in which he used gestures to indicate he was about to take the turn. Besides, it is important to bear in mind that some scenes containing interactive gestures may have disappeared from view after the final editing cuts.

Nevertheless, as previously mentioned, it is the very nature of our corpus that allowed for some interesting observations. The interviews occurred in settings that were informal—sometimes the communicators were walking on the street or having a conversation in a café or a bar. This fact allowed us to make new observations of gestures that had been described in literature, like the interactive gestures previously described by Bavelas et al.(1992), which had been recorded in a laboratory setting, with communicators sitting in a position facing each other. For the purposes those researchers aimed to achieve, that was an ideal setting; however, our corpus allowed us to observe these gestures in less than ideal circumstances, and allowed us to observe how these gestures were produced then.

As a result, one interesting finding is that there was a variation in the way the interactive gestures were produced, and yet they were recognizable. The turn passing gesture, for example, described in section 1.6.1, was produced differently while the communicators were walking down a street. As can be observed in Fig. 4, Pedro Andrade extends his hand forward. However, as his interviewee is walking by his side, this gesture does not point directly to him. Still, this gesture is easily recognizable, as will be discussed in the next section.

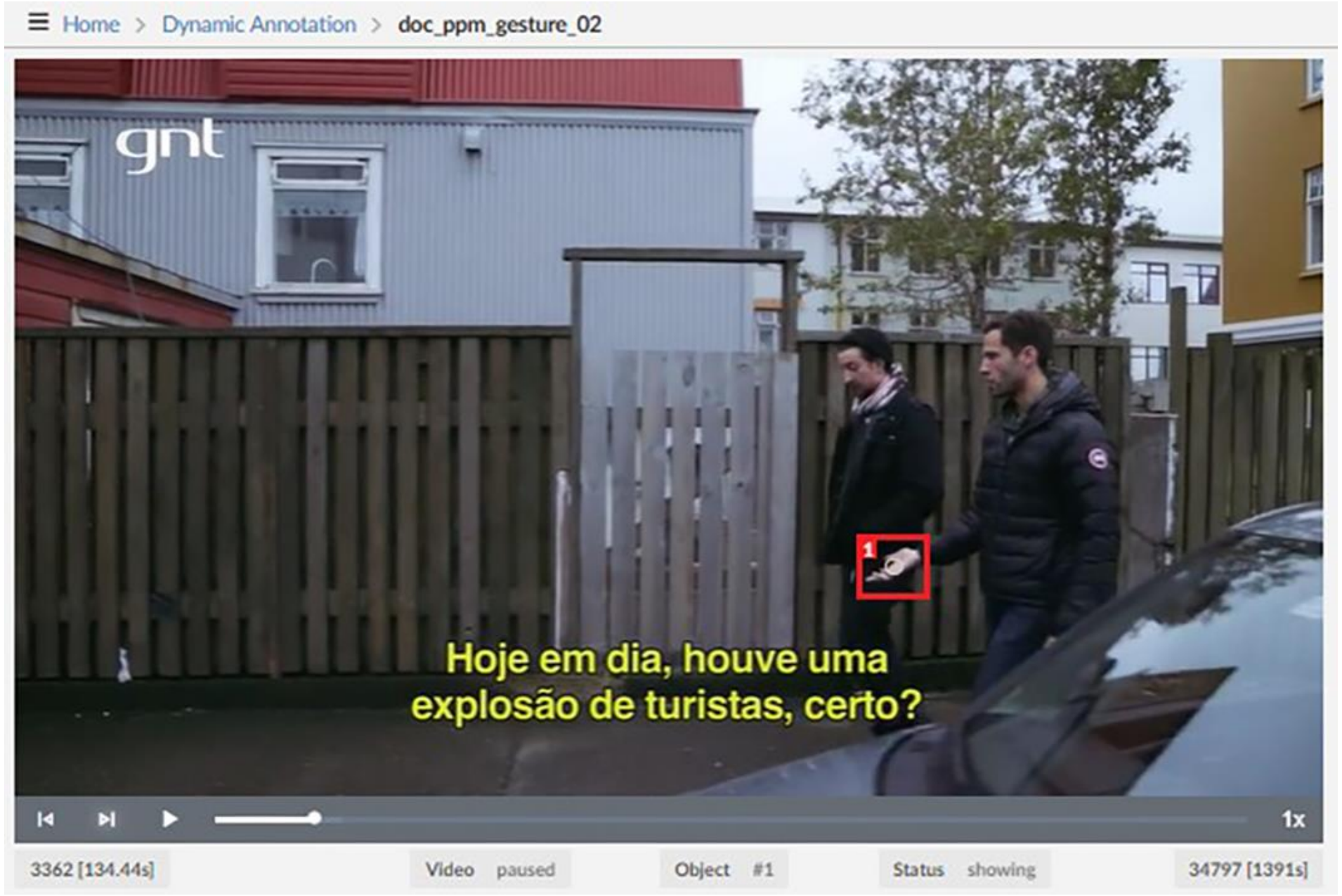

Figure 5 - Less prototypical turn passing gesture

Another gesture described by Bavelas et al. (1992) is the one in which the utterer requests help remembering a word. This is a gesture made with the hand producing circles in the air, with the hand moving between the directions of the utterer and the comprehender. During the same interview as the one shown in Fig. 5, Pedro Andrade's interviewee makes this gesture. However, as they were walking side by side, his arm is extended forward and the gesture is made sideways between the utterer and Pedro Andrade, as shown in Fig. 6. This gesture is also easily recognizable.

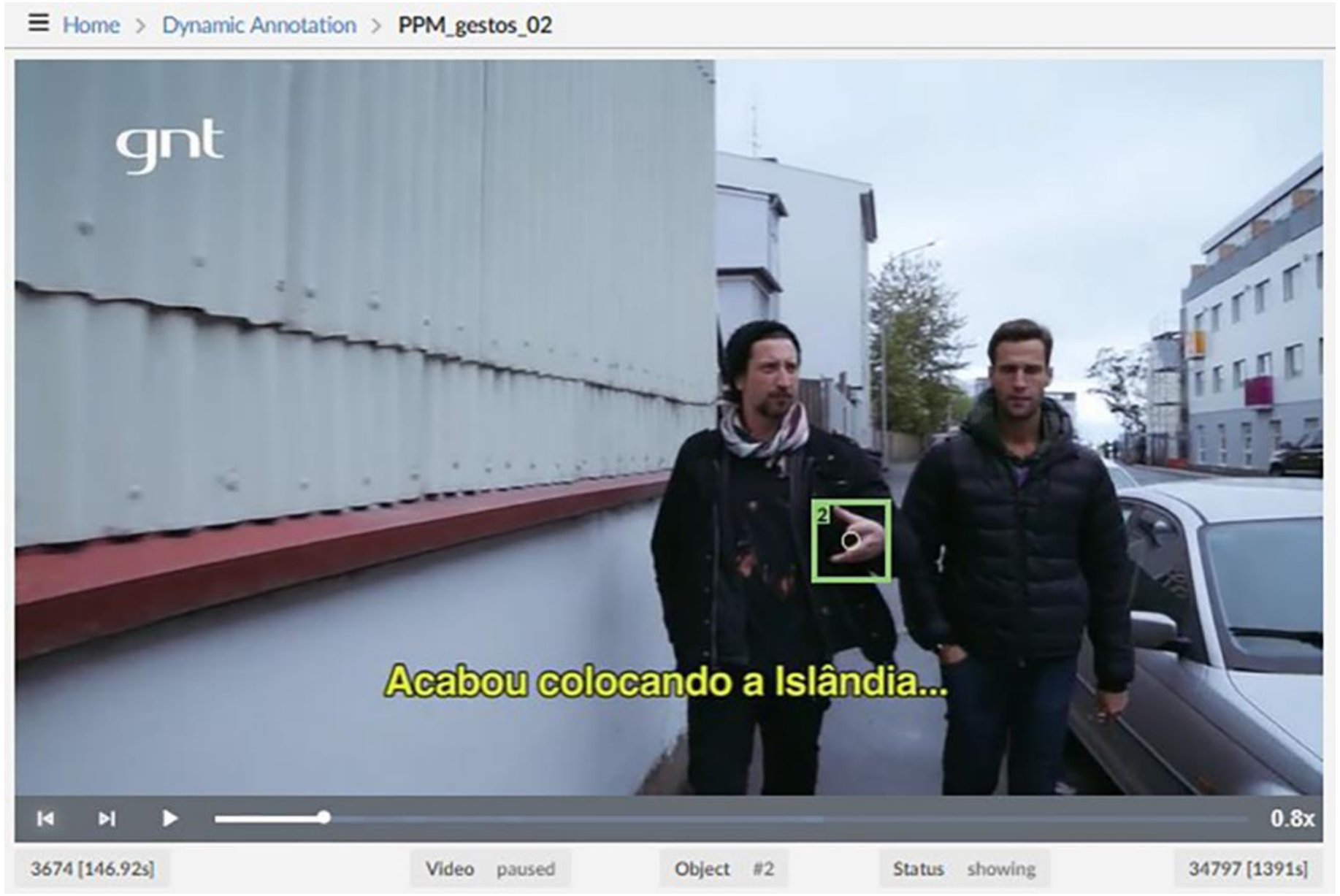

Figure 6 - Gesture indicating request for help remembering a word

A gesture that is not described in Bavelas et al. (1992) is the one for turn taking seen in Fig. 7. This gesture has a resemblance with the gesture of hands extending to take an object, and fulfills the criteria established by Bavelas et al. (1992) for recognition of interactive gestures: it could not be related to any of the words uttered at that moment, hands and face were in the direction of the comprehender, and it contained meaning that could be roughly paraphrased as, "let me take this (turn)". Moreover, Pedro Andrade's interviewee reacted to his gesture accordingly.

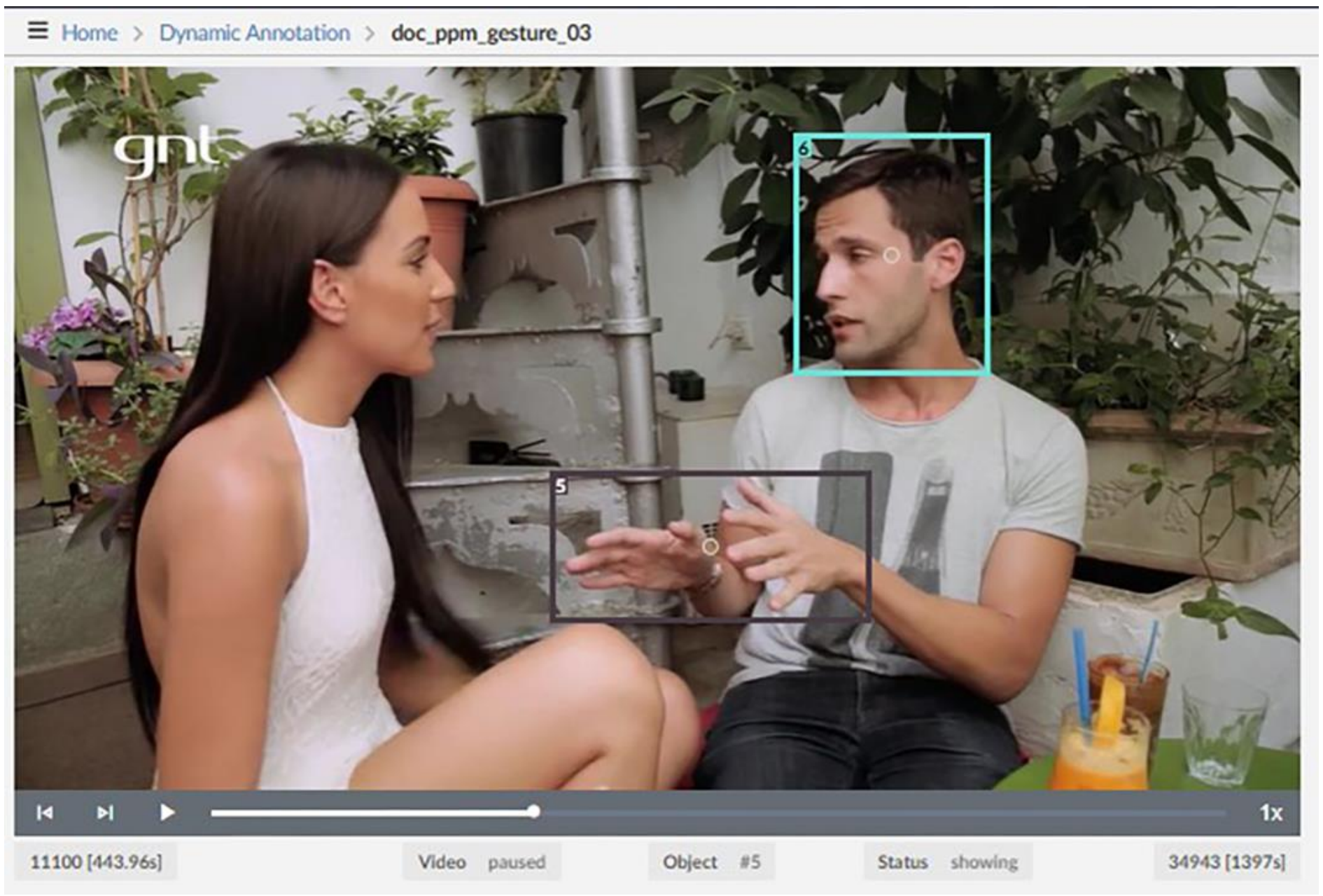

Figure 7 - Turn taking gesture

In summary, interactive gestures used in organization of conversation were observed through our corpus, which attested their usage and also demonstrated that some variations exist. The existence of these variations is explained in the next section.

## 4. DISCUSSION

As seen in sections 1.1 and 1.5, frames form the basis for mental spaces. On the other hand, processes of blending can lead to the conceptualization of frames. We propose that the conceptualization of pragmatic frames derive from processes of blending that involve Basic Communicative Spaces Network (BCSNs).

For a pragmatic frame to be invoked by the utterer, the process starts with the activation of the base space of the BCSN. Our conceptualization of the base space is, in itself, the result of blending that leads to the understanding of who plays the role of each communicator, in what space and time setting, which generates the understanding of how deictic terms such as ‘here’, ‘now’ and, more importantly, ‘I’ and ‘you’ are applied. One input space is structured by the frame for deixis and contains its elements, which are mapped to the elements in input 2, the space that contains the mental representation of the real people involved in a communicative situation. This mapping is what Fauconnier (1997, p. 120-121) calls a “Frame-to-values connection”. These elements, from the two input spaces, are projected into the blending space, which is the base space in the BCSN, that allows the communicators to recognize their roles as utterer, comprehender, and the place and time of communication. When the utterer says, ˜I˜, they are referring to themselves in the role of utterer, in the deictic center of communication. When the other person takes the conversational turn, that person also says, ˜I˜ to make the same reference.

Other mental spaces that are part of the BCSN are activated at the moment communication occurs, according to the context involving it. Take for example a situation when two acquaintances meet in the morning. Beyond the base mental space, another space is activated—the speech act space—which contains the social context that requires the two communicators to acknowledge each other’s presence and the fact that they are meeting for the first time within a certain part of the day. As Marmaridou (2000) explains, the social aspect of human beings cannot be ignored, as we exist in society. Whenever two people meet, they recognize the social structure they are inserted in. Moreover, social status, as Levinson (1983) explains, is conceptualized as proximity or distance between individuals, and even as position above or below. All these details are part of the knowledge of the world that is part of an adult’s conceptual system and is the foundation of the speech act space[11]. In our view, another blending process happens at this stage: the elements of the base space (input 1 of the new integration network) and the elements of the speech act space (input 2 of the new integration network), through conceptual integration, generates the blend that is the communicators as social beings, in a social context that requires a specific behavior, as seen in Fig. 8

[11] As seen in section 1.5, mental spaces are structured by frames or by general knowledge such as encyclopedic knowledge or knowledge of facts that occurred previously.

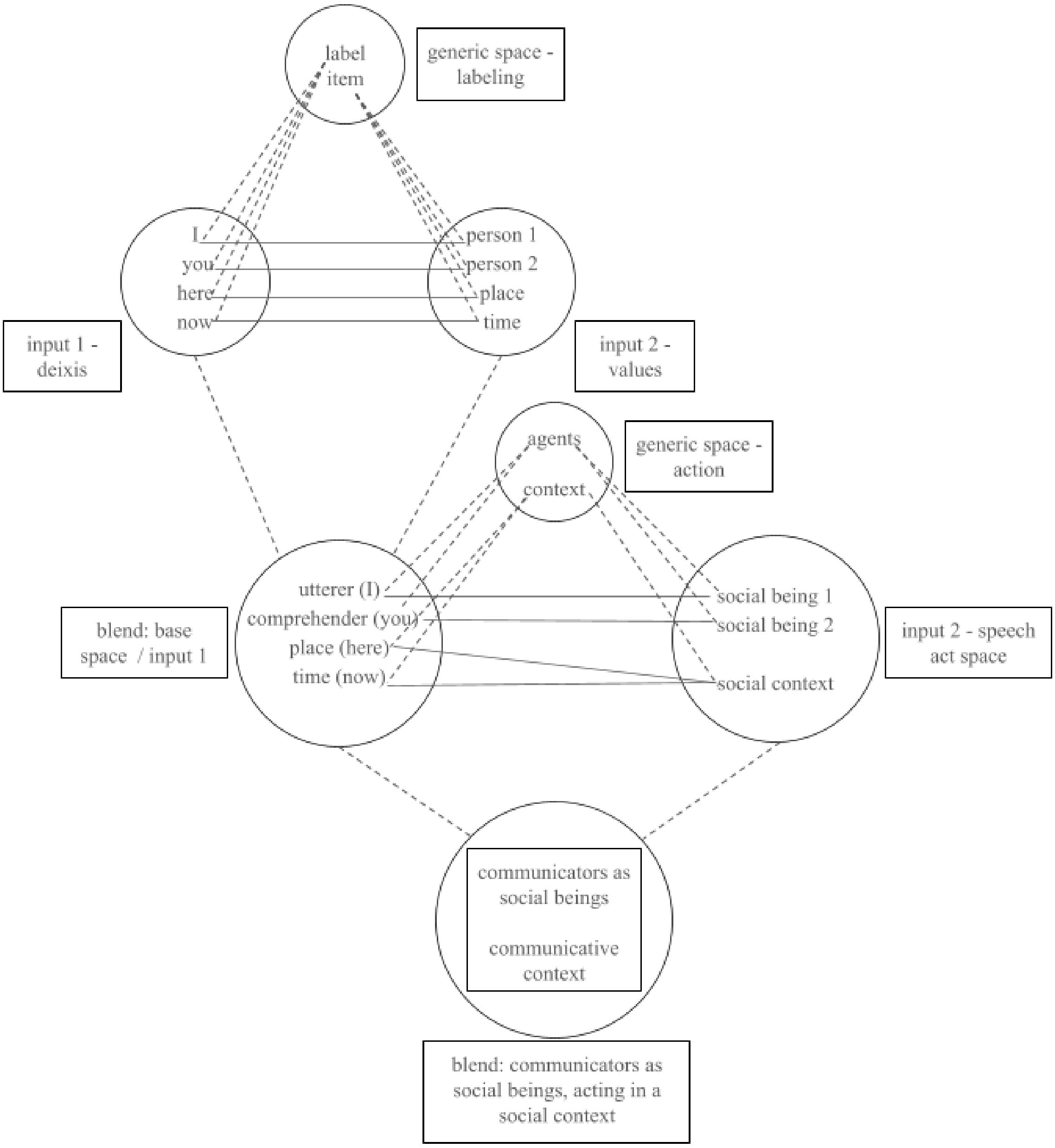

Figure 8 - Conceptual integration network

As an example of what it entails, let us consider a situation involving greetings: in Brazil, apart from informal expressions such as *oi* (‘hi’), there is a more formal term, *bom dia* (‘good morning’), used between midnight and noon. It is noteworthy that the literal meaning of bom dia is ‘good day’; however, the established custom is that this greeting is used only in the morning period, similar to the use of ‘good morning’ in English. The use of the term *bom dia* (‘good morning’), in opposition to a more informal *oi* (‘hi’) comes from the blending between the elements in the base space and the elements in the pragmatic space (the speech act space) that has as its basis the frame of what type of social encounter it is. The blending space, then, contains, as its elements, utterer and comprehender as social beings, with their perceived social positions. Once the type of social encounter is established, the metalinguistic space is activated, and the choice of the greeting, according to its level of formality, is made. The utterance chosen derives from another space that is then activated, which is a blending space resulting from all the conceptual integration processes above—the focus space.

This blending of mental spaces that can be explained through the BCSNs is not restricted, however, to the use of lexical terms. As explained by Fauconnier (1997) and Fauconnier and Turner (2002), processes of blending are part of mental processes present not only in language, but also in human cognition in general. After all, it is one of the tenets of cognitive semantics that language is part of general cognition. This is where the use of non-lexical pragmatic frames comes from, such as those evoked by interactive gestures in coordinating conversational turns.

Another important point in understanding the interactive gestures used in turn organization is the use of metaphors. During conversation, a speech turn is metaphorically conceptualized as an object, and as such, it may be passed from one speaker to the next and it may also be taken or kept. This conceptualization of speech turns is derived from the metaphor SPEECH TURN IS AN OBJECT, which, like so many other metaphors, is the result of blending[12]. One input space contains the elements of face-to-face conversation: the communicators, speech turns, subject of conversation, etc. The other input space contains people and also objects that can be manipulated by these people. The blendspace contains communicators as people who manipulate objects, and speech turns as objects that can be passed between these people or kept by them. The result of this blending process is that during conversation, the utterer makes a gesture as if they are passing a physical object when passing the turn, or a communicator who wishes to "take" the turn can make a gesture as if they are physically reaching out for an object.

As seen in section 4, there are cases in which the interactive gestures depart from their prototypes. We propose that the prototype theory by Rosch (1973) and Rosch & Lloyde (1978) accounts for these situations. For example, as explained above, when the utterer passes the turn, they tend to make a gesture that represents, in its most prototypical use, the passing of an object to the person chosen by them to be the next speaker. This prototype is the basis through which we are able to recognize less prototypical gestures as part of the category "passing the speaking turn". This way, a gesture does not need to be exactly the same as described previously in literature for it to be recognized by communicators and evoke a specific pragmatic frame.

## CONCLUSIONS

Our work has demonstrated that a cognitive linguistics approach can allow us to better understand the processes through which human cognition gives rise to meaning in the form of pragmatic frames, and how these frames can be evoked as a tool for face-to-face conversation turns to be organized by communicators.

As demonstrated, FrameNet Brasil has developed an annotation methodology for pragmatic frames and has proven that these annotations enriched the Frame$^2$ dataset by providing additional information to the studies of multimodality within the field of frame semantics. These annotations confirm the conclusions reached by Czulo, Ziem and Torrent (2020) and the fact that pragmatic frames are evoked by verbal communication as well as by specific gestures.

Moreover, this corpus represents an opportunity for observation of interactive gestures in situations which are less controlled than the conditions that could be expected within a laboratory, demonstrating that some of these gestures can be different from what had been previously recorded. We expect that the future development of a larger dataset can provide us with more information on interactive gestures and pragmatic frames used for organization of conversation turns.

## ACKNOWLEDGEMENTS

This paper is a project within ReINVenTA – Research and Innovation Network for Vision and Text Analysis of Multimodal Objects. ReINVenTA is funded by FAPEMIG grant RED 00106/21, and CNPq grants 408269/2021-9 and 420945/2022-9. Helen de Andrade Abreu's research is funded by Minas Gerais State Agency for Research and Development (FAPEMIG) grant RED-00106-21, code 41330. Tiago Timponi Torrent's research is funded by National Council for Scientific and Technological Development (CNPq) grant 311241/2025-5.

[12] For more on metaphors and blending, see Fauconnier and Turner (2002) and Dancygier and Sweetser (2014).